# Incorporating Large Language Models into Production Systems for Enhanced Task Automation and Flexibility
*Moving Towards Autonomous Systems*

M.Sc. **Yuchen Xia**, B.Sc. **Jize Zhang**, Dr.-Ing. **Nasser Jazdi**, Prof. Dr.-Ing. Dr. h. c. **Michael Weyrich**
yuchen.xia@ias.uni-stuttgart.de, st171260@stud.uni-stuttgart.de, nasser.jazdi@ias.uni-stuttgart.de, michael.weyrich@ias.uni-stuttgart.de
Institute of Industrial Automation and Software Engineering, University of Stuttgart

**Abstract**
This paper introduces a novel approach to integrating large language model (LLM) agents into automated production systems, aimed at enhancing task automation and flexibility. We organize production operations within a hierarchical framework based on the automation pyramid. Atomic operation functionalities are modeled as microservices, which are executed through interface invocation within a dedicated digital twin system. This allows for a scalable and flexible foundation for orchestrating production processes. In this digital twin system, low-level, hardware-specific data is semantically enriched and made interpretable for LLMs for production planning and control tasks. Large language model agents are systematically prompted to interpret these production-specific data and knowledge. Upon receiving a user request or identifying a triggering event, the LLM agents generate a process plan. This plan is then decomposed into a series of atomic operations, executed as microservices within the real-world automation system. We implement this overall approach on an automated modular production facility at our laboratory, demonstrating how the LLMs can handle production planning and control tasks through a concrete case study. This results in an intuitive production facility with higher levels of task automation and flexibility. Finally, we reveal the several limitations in realizing the full potential of the large language models in autonomous systems and point out promising benefits. Demos of this series of ongoing research series can be accessed at: https://github.com/YuchenXia/GPT4IndustrialAutomation

1. Introduction
**Automation systems vs. Autonomous systems**

In the evolving landscape of technology, the terms automation and autonomous systems are often intertwined yet distinctly different in impact. Both involve the use of technology to perform tasks with minimal or no human intervention, but differ significantly in their flexibility in decision-making:

- **Automation**: Traditional automated systems generally follow rigid, predefined rules and workflows and are not designed to adapt to changes unless those changes have been anticipated and programmed into the system. A usual pre-requisite of automation lies in repeatability and predictability, which limits its adaptability to dynamic environments.

- **Autonomy**[1]**:** on the other hand, autonomy entails a higher level of adaptability and decision-making capability. An autonomous system can adapt to un-predefined changes and utilize knowledge to make decisions based on available options and system objectives, demonstrating flexibility in problem-solving and task execution, whereas automation typically does not. This concept is commonly applied in fields such as intelligent robotics and artificial intelligence, and it is also used in political contexts to describe self-determined individuals or systems that make choices aiming to achieve optimal outcomes.[1]

In the transition from automation to autonomy, the key differentiator is intelligence—the capability to make informed, dynamic decisions. This intelligence cannot be practically provided by using exhaustive rules due to the unpredictability and variability of real-world environments. Such rules cannot cover every possible scenario and often struggle with granularity—they can be either too broad, failing to address specific situations, or overly detailed, making them cumbersome to exhaustively implement and maintain. Moreover, maintaining and updating such a comprehensive rule set demands extensive engineering effort. The technological development in natural language understanding indicates the superiority of machine learning with neural networks over rule-based systems, as the top solutions are all based on neural networks [2].

Large language models (LLMs) can offer the intelligence to bridge the gap between traditional automation and autonomy in industrial systems. These models internalize the knowledge patterns learned from training data conveying collective human knowledge, and they are capable of interpreting complex text and performing dynamic reasoning based on the given input. Their general adaptability allows them to respond to new situations and conditions without the need for specific re-training.

Incorporating LLMs into industrial automation systems empowers us to utilize their capabilities in performing diverse tasks within industrial automation, further reducing the need for human intervention in tasks that require intelligence. LLMs are particularly effective in extracting important information from vast datasets, comprehending texts, making reasoning, solving problems, and supporting decision-making processes [3]. By enabling these models to perform intelligent information processing based on real-time data, they can swiftly adapt to changes, thereby boosting efficiency and productivity.

In the following sections, we explore the integration of LLMs into automated production systems to enhance their autonomy and flexibility. Section 2 reviews the typical state-of-the-art frameworks of industrial automation systems, establishing the background of our approach. Section 3 introduces a novel design framework that effectively integrates LLMs for dynamic process planning and control in production systems. Sections 4 and 5 present a pilot case study on our modular production facility, demonstrating the practical effect of our LLM-integrated approach. Sections 6 and 7 discuss the evaluation, its implication and the current limitations. Section 8 summarizes the the result and the outlook for implementing LLMs in autonomous systems.

---

[1] 'Autonomous' derives from the Greek roots: 'auto,' meaning 'self,' and 'nomos,' meaning 'custom' or 'law.'

## 2. Background
### 2.1 Automation pyramid

To manage the diverse tasks in industrial automation, the automation pyramid is a conceptual framework organizing the tasks in industrial automation systems into five distinct layers [4]: starting at the base, the **Field Level** involves direct interaction with physical technical processes through sensors and actuators. Above this is the **Control Level**, where devices like programmable logic controllers (PLCs) execute real-time control tasks based on functional execution of control and regulation. The **Supervisory Level** includes systems such as SCADA and human-machine interfaces (HMI), enabling operators to monitor and adjust processes. The **Production Planning Level** manages optimized production planning and scheduling, coordinating operational logistics and production processes. At the top, the **Enterprise Level** integrates various business functions through enterprise resource planning (ERP) systems, enhancing decision-making and resource management across the organization. This hierarchical model ensures systematic data flow and task execution across different facets of industrial operations.

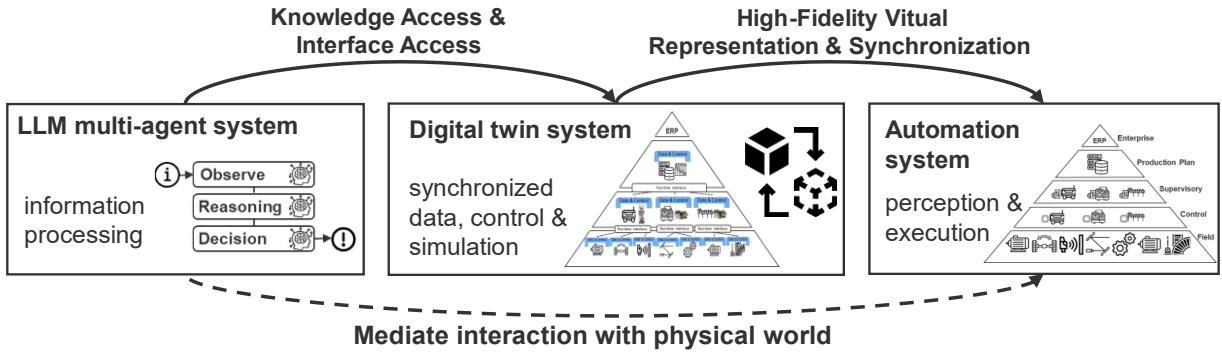

Figure 1 Autonomous system enabled with LLM multi-agent system, digital twins and automation system

### 2.2 Digital twins for automation system

Moving beyond traditional automation, the integration of digital twins facilitates data-driven optimization and enables remote interactions. This integration creates a software system synchronized with physical systems. As a result, operational changes are immediately reflected, providing precise control interfaces for effective process management.[5] The digital twins can serve as the connecting bridge between LLM and the automation system, as shown in Figure 1.

**2.3 LLM as an enabler of intelligent autonomous system**

The incorporation of LLMs into the system enriches the digital twin with the intelligence for advanced data analysis and decision-making. These models leverage vast amounts of text to perform reasoning, provide insights, solve tasks, and support data-driven decision-making, providing general intelligence to a wide range of problems. This intelligence can be utilized to enhance adaptability and flexibility within the automation system, leading to quicker reactions to changes and automated problem-solving.

**3. Methodology**

**3.1 LLM agent and task decomposition for LLM multi-agent system**

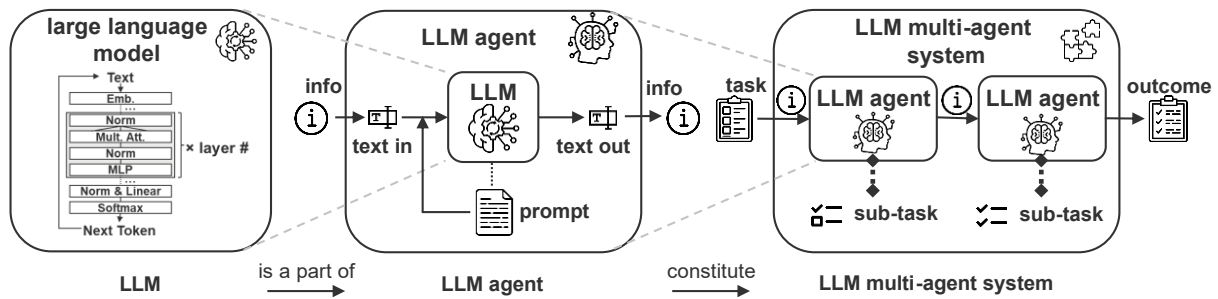

Figure 2 the general method to create a LLM agent based on LLM and its integration into a multi-agent system for complex problem-solving.

At the core, a generative LLM is a deep neural network for processing and generating text-based information, as illustrated in Figure 2. This capability allows them to act as intelligent agents, equipped to interpret diverse information and generate output as a response to a task. In the design of an LLM agent, the prompt is crucial for directing the LLM's behavior to a specific task context, and this mechanism is also referred to as "in-context learning" or "prompt learning"[6]. The input text for a task is incorporated into prompt templates, which guide the LLM in generating appropriate outputs, thereby enabling the LLM agent to function effectively as an information processor. An example is shown in Appendix A.

If a task is too complex for a single LLM agent to handle, task decomposition becomes necessary. This is where the design of LLM agents as integral components of a multi-agent system comes into play. Each LLM agent is tasked with solving specific sub-tasks within the overall task-solving process, as shown in Figure 2. These sub-tasks can range from interpreting data inputs and generating dynamic process plans to determining the next executable microservices as actions. The agents operate within a framework where complex tasks are broken down into more manageable sub-tasks, allowing individual agents to execute them more effectively.

## 3.2 Structured prompt and agent design template

To develop an effective prompt for a LLM to perform a specific task, it is essential to address several key aspects to guide the text generation behavior. Drawing from our experience in creating LLM-based applications [7], [8], [9], we outline these aspects in a template for effective prompting. This template includes the following critical elements:

- **Role, responsibility and goal** outline the agent's functional role and the objectives it aims to achieve, ensuring alignment with its responsibility and broader system goals.
- **Context specification** supplies detailed background information relevant to the task, enabling the agent to interpret the task-specific context.
- **Behavioral instructions and constraints** contain specifications to guide the agent's responses, ensuring they adhere to task responsibility.
- **Input-output abstract pattern** defines the abstract formats for input and output, which helps standardize the agent's interaction and generate parsable output.
- **Input-output interaction field** provides dynamic input and leaves the output section with only a cue word "*Output:*", forcing LLM only generate output content by text continuation. (For instance, c.f. Appendix A)

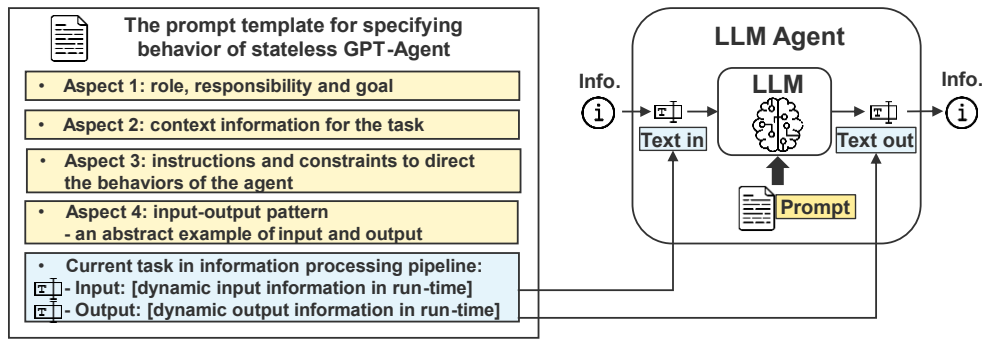

Figure 3 Prompt template for creating an effective LLM agent for information processing task

This template offers a systematic structure for specifying prompts, facilitating the effective and reproducible use of LLMs to develop software components with clearly defined interfaces (i.e., LLM agents).

## 3.3 Design of dedicated LLM agents for tasks on different layers in automation system

Building on the foundational methods previously introduced, a multi-agent system can be designed to perform diverse tasks across different layers of the automation pyramid.

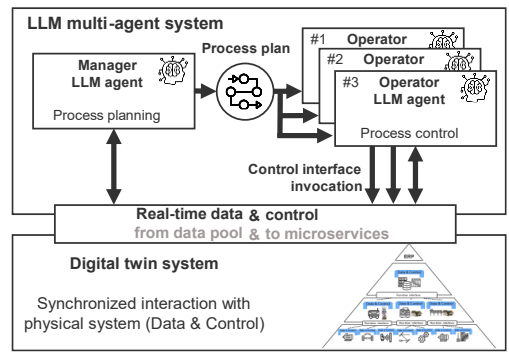

Figure 4 Building LLM multi-agent system on the foundation of digital twin system

Within the LLM multi-agent system, each LLM agent is assigned a specific role. At the planning level, a manager agent is responsible for the overall planning of the production operations. At the control level, operator agents are responsible for the control tasks, making decisions based on real-time data to decide actionable operations as microservice interface calls.

The digital twin system is the foundation for the multi-agent system by providing real-time information of the physical processes and control interface. Field-level executions are managed as microservices, which can be invoked via a run-time interface when required. This integration enables LLM agents to have immediate access to data and control mechanisms through the digital twins' software interfaces, as shown in Figure 4.

## 4. Case study setup

At the Institute of Industrial Automation and Software Engineering (IAS) at the University of Stuttgart, our laboratory is equipped with a modular production system, as illustrated in Figure 5. This system features several automation modules capable of executing a range of production and logistics tasks. We have retrofitted the system to support real-time data access and control interface through a run-time environment [7].

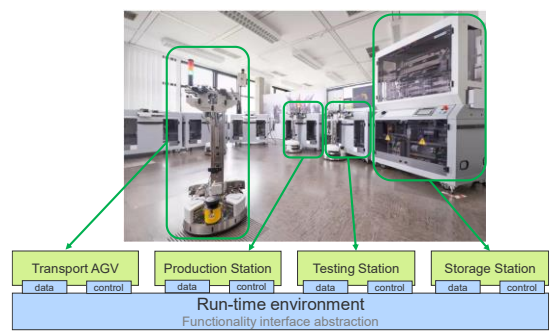

Figure 5 Cyber-physical modular production system at IAS, University of Stuttgart

Following the methods described in the last sections, we implemented a digital twin system and built an LLM multi-agent system. The LLM agents are designed on the planning level and the control level according to the automation pyramid, as illustrated in Figure 6.

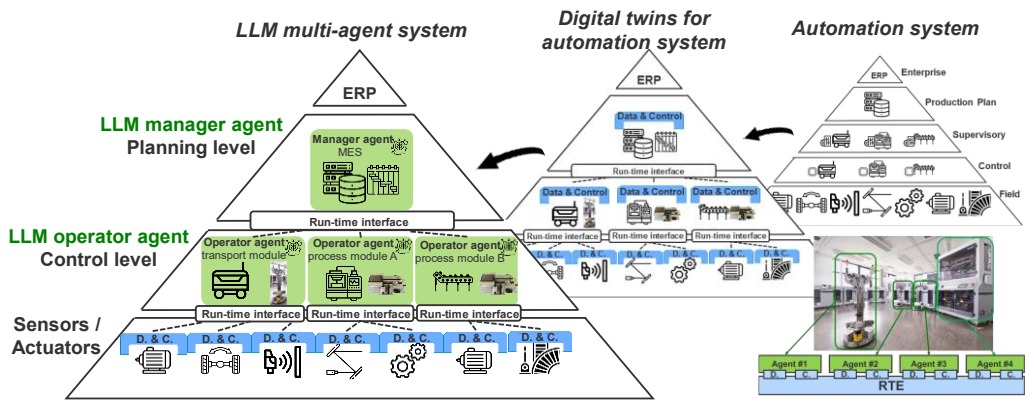

Figure 6 Upgrade the automation with digital twins and LLM multi-agent system

## 5. System design
### 5.1 Conceptual design

In our designed system, LLM agents play a central role by interpreting production-specific information provided by digital twins, planning operations, and making decisions. They respond to user commands or system-triggered events, which can be entered through an application front-end or detected through continuous monitoring.

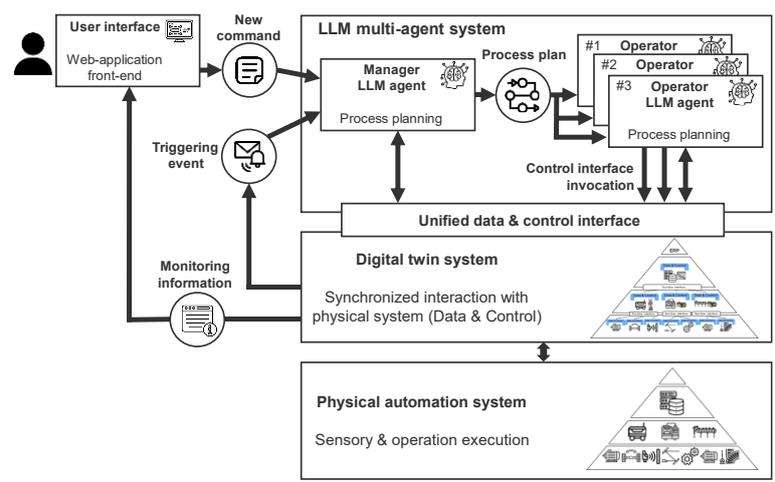

Figure 7 The conceptual system design of LLM enhanced autonomous production system

Once a task-triggering event is detected, the manager LLM agent dynamically generates a process plan consisting of a sequence of subtasks for operator agents. The operator agents then break this plan down into atomic operations organized as microservices, each encapsulating a specific, executable operation as a part of the production process. This decomposability of the task plan and microservices ensures modularity, scalability, and flexible orchestration of operations.

## 5.2 System components

In Figure 8, a more detailed explanation of the system components and their interactions is depicted.

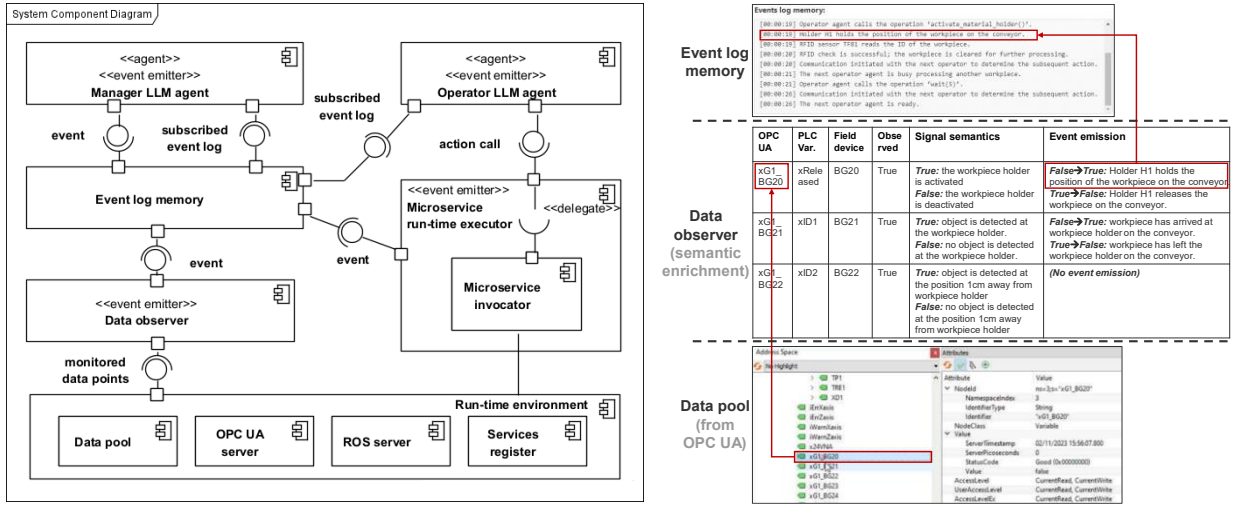

Figure 8: System components diagram (left) and the semantic enhancement process of the information from data pool, data observer to event log memory (right).

### 5.2.1 Run-time Environment

As shown in Figure 8, the run-time environment includes the following software components:

- **OPC UA server:** The OPC UA server interfaces with PLCs that control the production modules' functionalities. It reads and sets values to access and modify the operational parameters and statuses of PLCs. The functionalities programmed into the PLCs, such as control logic for various production operations, are encapsulated into atomic services. These services, such as "conveyor_run(direction, duration)" and "activate_material_holder()", are designed to be independently executable.

- **ROS server:** In parallel, the ROS Server is essential for the control and coordination of AGVs for the logistics functionalities. Functionalities such as "move_to(station_ID)" and "load_workpiece()" are designed to control AGV actions.

- **Service register:** This component contains a lookup table and service interface descriptions, cataloging all available microservices. This component facilitates the discovery and the invocation of services as needed. The microservice architecture allows for the independent implementation and scaling of these atomic services. This modularity enhances the system's operational compatibility and adaptability, making it possible to handle a variety of operational scenarios by orchestrating atomic functionalities into a process sequence. An excerpt of the microservices for an operator of a conveyor transport agent is shown in Table 1 for example.

- **Data pool:** This is a centralized repository that aggregates real-time operational data from diverse sources of the system and provides data access to external components.

Table 1 Excerpt of the listed microservices in the component "*Microservice Register*"

| Micro-services | Description |
| --- | --- |
| conveyor_belt_run(direction, duration) | Moves the conveyor belt in the specified direction ('forward' or 'backward') for a set duration in seconds. |
| conveyor_belt_stop() | Stops the conveyor belt. |
| activate_material_holder() | Engages a mechanism to hold the workpiece in place on the end of the conveyor. |
| deactivate_material_holder() | Releases the holding mechanism, freeing the workpiece from the secured position at the end of the conveyor. |
| communicate_with_agent(agent_id) | Send a message to the next agent in the production line to coordinate the handover or next steps. |
| release_workpiece_to_next_agent() | Release the workpiece at the ready position to the next agent and transfer the control of this workpiece to the next agent. |
| wait(duration) | Pauses the current operation for a set duration in seconds. |
| send_alert_to_human_supervisor(issue) | Alerts a human supervisor about issues. |
| pass() | Executes no operation, allowing the system to bypass this command without making any change. |
| … | … |

The semantic enrichment process is pivotal for transforming technical operational data from the ***Data pool*** into semantically rich information that can be used for higher-level planning and control tasks in ***Event log memory***. The process is detailed in Figure 8 (right), in which the arrows illustrate how the hardware-level data are semantically enriched and translated into actionable insights for the system's agents. In the middle of this process, the ***Data observer*** monitors changes in the data and applies specified rules to detect and contextualize these changes into significant events necessary for planning and control. These enriched events are then logged in the ***Event log memory***, providing a structured information basis that LLM agents utilize for reasoning and dynamic decision-making.

### 5.2.3 Event-driven information modeling in text

Several components in Figure 8 are typed as ***event emitter***. They can send event information to an ***Event log memory***, which organizes events in chronological order. Each event in the log is tagged with a standardized system timestamp, ensuring that information about changes, decisions and actions can be captured and further consumed by each LLM agent according to its interest scope within a subscription mechanism. In the ***Event log memory***, an event is captured at timeless moments without a duration: instead of recording a process "the material holder secures the position of the workpiece for 5 seconds", it is logged in two separate entries: "[00:00:00] the material holder holds the workpiece" and "[00:00:05] the material holder releases the workpiece."

The necessity for this event-driven design arises from the nature of planning and control problems, in which the dimension of time is indispensable. However, invoking an LLM for decision-making is a discrete action rather than a continuous process. This characteristic necessitates an approach to structure time-aware information in the form of the event log. Based on this design, an excerpt of the event log containing the necessary atomic information (event) can be integrated into the prompt of an LLM agent for decision-making.

### 5.2.4 Decision-making by LLM agents and action execution with microservices

The **LLM agents** subscribe to specific events within the **Event log memory**. Based on the events logged, the agent makes informed decisions about the production planning and control actions depending on the dynamic situation. The decisions generated by the LLM are structured as textual data in JSON format, which can easily be handled with code script. The JSON data is parsed into actionable function calls. These are then passed either to another **LLM operator agent** for further handling or directly to the **Microservice run-time executor**, which carries out the commands. The process is depicted in the following sequence diagram in Figure 9 for instance.

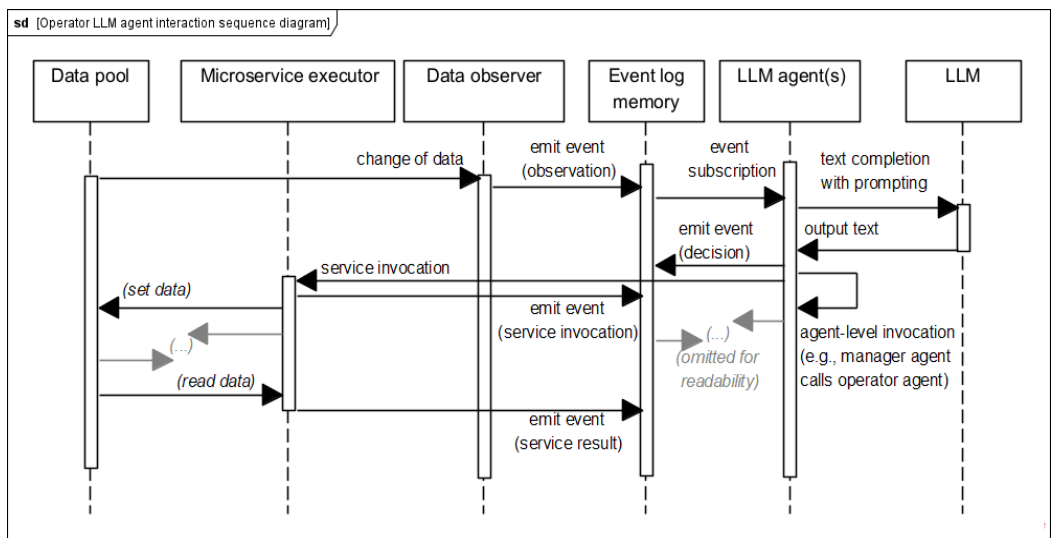

Figure 9 Sequence diagram of the interaction between system components from data change to action execution.

### 6. Result and Evaluation

The system has been implemented in our laboratory. It operates as follows: when a user inputs a task via the front-end user interface or a triggering event is identified, the LLM manager agent formulates a production plan, as shown in Figure 10.

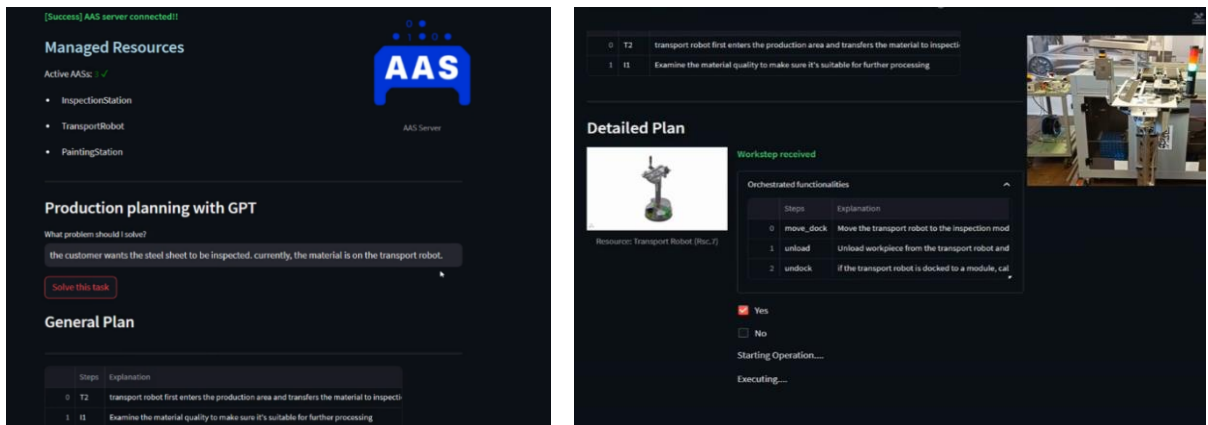

Figure 10 The user interface and the live demonstration of operation planning and control

The operation steps in this plan are subsequently delegated to operator agents who arrange a sequence of atomic microservices to execute the task. This prototype successfully demonstrates the concept of integrating LLMs into an automated production system, moving towards autonomous systems.

The evaluation of the developed prototype is outlined in the following sections across three key aspects. Further details are available in our GitHub Repository [2], as we continue comprehensive testing and evaluation of this approach.

**Evaluation of different LLMs for planning and controlling capabilities**

To assess the foundational automation capabilities of our system, we evaluated how accurately the LLMs could interpret basic tasks provided via prompts and subsequently execute the correct operations. These operations include processing routine orders, adjusting settings according to user requests, managing standard material transfers, and reacting to unforeseen and un-preprogrammed events that arise during operation. We applied different LLMs for powering the LLM agents for our system design, including proprietary model "GPT-4o" and GPT-3.5, and two opensource models "LLAMA-3-8B", "LLAMA-3-70B", as well as a fine-tuned model based on "LLAMA-3-70B", which is trained on the texts in the event logs during the system operation.

We selected a total of 100 test points to evaluate the performance of the language model agents. These points were divided into two categories: 50% comprised predefined routine planning and control tasks, for which there are direct instructions in the prompts on how to execute the tasks in a standardized way (standard operation procedure), and the remaining 50% consisted of scenarios that were not pre-defined in prompts, requiring the LLM agents to generate spontaneous decisions.

---

[2] A demo and the detailed evaluation results of this series of research are released at the Github-Repository: https://github.com/YuchenXia/GPT4IndustrialAutomation

The primary focus of our evaluation was to assess the system's capability to autonomously handle both pre-defined and unforeseen events that occur during operation. For example, a testing scenario is defined as follows: a workpiece being transported on a conveyor fails to reach its next designated position after the activation of conveyor for 10 seconds. In this case, the operator agent initially decides to wait for the workpiece's arrival. However, once the conveyor automatically stops, the LLM agent system takes proactive measures by not only alerting the human supervisor but also suggesting reactivating the conveyor belt. This action exemplifies a response that goes beyond the pre-defined prompts or descriptions in the microservices.

The evaluation was structured on two critical levels:
- The system's ability to generate error-free executable commands.
- The effectiveness of these commands in addressing and resolving the incidents encountered, as judged by human evaluators.

To illustrate, an error-free executable command like "wait(5)" would be generated if the workpiece does not move even if the conveyor is activated. Although technically correct, this command is not deemed effective or optimal. The more appropriate response, which exemplifies both error-free execution and effective problem-solving, would be "send_alert_to_human_supervisor(issue)".

The evaluation results are shown in Figure 11.

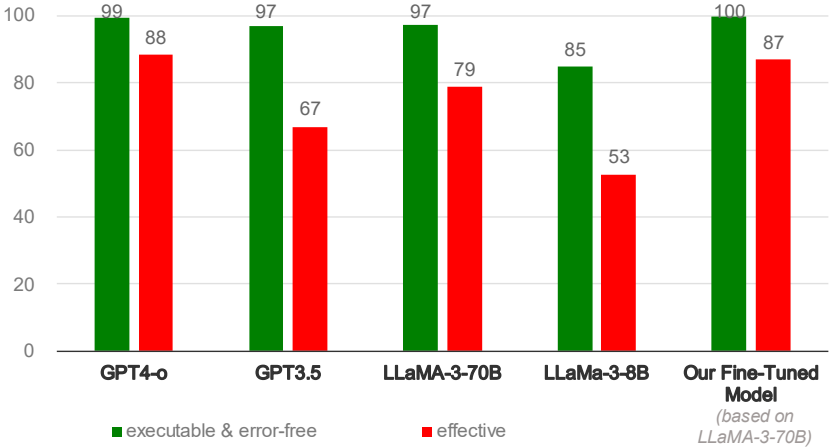

Figure 11 Evaluation of planning and control capabilities across various llms for the designed system

The effectiveness of the commands varies significantly among the models. While "GPT-4o" and our fine-tuned model based on LLAMA-3-70B score similarly high (88% and 87% respectively), this confirms the effectiveness of fine-tuning for this specific task. "LLAMA-3-8B" shows a notable dip to just 53%, indicating that smaller models may not be capable of performing complex planning and control tasks effectively.

## 7. Limitation

### 7.1 Real-time Performance

While the system integrates LLM agents effectively for planning and executing tasks, real-time performance remains a challenging aspect. The latency introduced by processing and generating responses through LLMs can impact the system's ability to operate in strict real-time environments. Due to this limitation, in our system, LLM reasoning is only performed when the operations can be paused or when new events during the LLM inferencing do not affect the decision-making. The LLM inferencing typically lasts between 100 ms to 1 second when utilizing dual NVIDIA A100 Tensor Core GPUs deployed on a local server, and 1 to 10 seconds for cloud server. For this reason, the generated text should remain concise by design.

### 7.2 Comprehensive Testing

The complexity and variability of production environments necessitate extensive validation of any automated system to ensure its robustness and reliability. Our current evaluation primarily relies on synthetic test cases designed to simulate typical scenarios within the controlled environment of our laboratory. These are used for initial proof-of-concept demonstrations, and the test cases could quickly become complicated and condition-specific. The system-generated decisions and operations still require human approval, making it only sufficient for an assistant system. Moving forward, we will continue to perform more comprehensive testing of the system to better address the dynamic nature of real-world manufacturing processes. It will be crucial to employ test-driven development methods to identify and mitigate potential failures or inefficiencies, thereby delivering greater value and enhanced usability.

### 7.3 Cost-Benefit Evaluation

The economic feasibility of integrating LLMs into production systems remains an area requiring thorough investigation. While some theoretical benefits are evident, such as more intuitive human-machine interaction, reduction of human effort, training cost and reaction time to changes, the actual cost related to their implementation, improvement, and maintenance are not yet fully understood. The benefits of increased adaptability, automation efficiency, and productivity gains can be further quantitatively assessed against these costs, in order to validate the long-term economic viability of deploying large language models in industrial settings.

## 8. Conclusion

This paper presented a novel approach to integrating LLMs into industrial automation systems, thereby enhancing task automation and flexibility. By employing LLMs within a multi-agent system and integrating them with digital twins and automation systems, we demonstrated their efficacy in autonomously managing production planning and control tasks. The integration of LLMs offers significant advantages, including dynamic decision-making and planning, as well as creative problem-solving for un-predefined issues, which can be readily developed as intelligent assistant systems. This integration significantly reduces the need for human monitoring and trouble-shooting, as well as decreases reaction times to un-predefined changes. Our results show that LLMs can effectively handle both routine and unexpected events, thereby improving operational efficiency and adaptability. A fine-tuned model based on an open-source model can have equivalent performance to proprietary GPT-4 on this specific application task.

However, challenges such as real-time performance latency, the need for comprehensive real-world testing, and the economic implications of deploying LLMs remain. These issues highlight critical areas for further research and optimization to fully harness the potential of LLMs in industrial settings.

In summary, integrating LLMs into production systems offers promising prospects for advancing toward intelligent autonomous systems. Drawing on insights from experience, we hold the view that developing future intelligent autonomous systems powered by LLMs will require a test-intensive or even test-driven development process. Our future research will focus on broadening application areas and improving the system through systematic testing and evaluation. This will enhance the value and cost-effectiveness of these systems, continuing to realize the transformative impact of LLMs on industrial automation.

**Acknowledgements**

This work was supported by Stiftung der Deutschen Wirtschaft (SDW) and the Ministry of Science, Research and the Arts of the State of Baden-Wuerttemberg within the support of the projects of the Exzellenzinitiative II.

## Appendix A

## Prompt for an operator agent: (1025 tokens)

You are an operator agent that is responsible for controlling the material transport on a conveyor before a production process.

This conveyor belt is a straight, 1-meter-long system designed for material transport. At its entrance, sensor BG56 detects incoming workpieces. At the end of its path, sensor BG51 detects the workpiece at the ready position, actuator holder H1 can secure the workpieces in place, and an RFID sensor TF81 reads the workpiece IDs for processing validation.

Components descriptions:
Sensors:
BG51: Detects workpieces at the ready position.
RFID Sensor TF81: Reads workpiece IDs to validate processing criteria.
Actuators:
Conveyor C1: Controls the movement of the conveyor. It can be controlled via the following command(s):
conveyor_belt_run(direction, duration_in_second), conveyor_belt_stop().
Material Holder H1: Holds workpieces at the ready position. It can be controlled via the following command(s):
activate_material_holder(), deactivate_material_holder().
Actions you can take:
conveyor_belt_run(direction, duration_in_second): Moves the conveyor belt in the specified direction ('forward' or 'backward') for a set duration.
conveyor_belt_stop(): Stops the conveyor belt.
activate_material_holder(): Engages a mechanism to hold the workpiece in place on the end of the conveyor.
deactivate_material_holder(): Releases the holding mechanism, freeing the workpiece from the secured position at the end of the conveyor.
communicate_with_next_agent(): Send a message to the next agent in the production line to coordinate the handover or next steps.
release_ready_workpiece_to_next_agent(): release the workpiece at the ready position to the next agent and transfer the control of this workpiece to the next agent.
wait(duration_in_second): Pauses the current operation for a set duration in seconds.
send_alert_to_human_supervisor(): Alerts a human supervisor about issues.
pass(): Executes no operation, allowing the system to bypass this command without making any change

Standard Operation Procedure:
The process begin with a workpiece arriving at the entrance of the conveyor.
1. If sensor BG56 detect an object, it indicates that a workpiece is detected at the entrance position. You should call activate_conveyor(forward, 10) to set the conveyor moving forward for 10 seconds, to transport the workpiece to the ready position.
2. If sensor BG51 detect an object, it indicates that a workpiece is detected at the ready position. You should call activate_material_holder() to secure the workpiece in place, ensuring that the workpiece is securely positioned.
3. If the workpiece is detected at the ready position and is being held, you should call rfid_read() to read the workpiece information, to determine whether the workpiece is suitable for further processing.
4. If the workpiece information checks out OK, you should call ask_next_operator() to determine the status of the next operator agent, in order to decide whether to wait or to forward the workpiece to the next operator agent.
5. If the next operator is busy, then call wait(5) to wait for 5 seconds before calling ask_next_operator() again to check the status of the next agent; if the next operator is ready, then call release_workpiece() to release the workpiece and hand it over to the next operator.

Instructions for you:
You will observe an event log in the following input section, and you shall generate your response in the output section.
You should follow this input and output pattern to generate a response in JSON format:
Input:
// An event log will be given here.
Output:
{"reason": "a_reason", "command" : "a_command()"}

Now, you should generate a response:
Input:
[00:00:14] Sensor BG56 detects an object at the entrance.
[00:00:14] Operator agent calls the operation 'conveyor_belt_run(forward, 10)'.
[00:00:14] The conveyor starts moving forward.
[00:00:19] Sensor BG51 at the ready position detects the workpiece.
[00:00:19] Operator agent calls the operation 'activate_material_holder()'.
[00:00:19] Holder H1 secures the position of the workpiece on the conveyor.
[00:00:19] RFID sensor TF81 reads the ID of the workpiece.
[00:00:20] RFID check is successful; the workpiece is cleared for further processing.
[00:00:20] Communication initiated with the next operator to determine the subsequent action.
[00:00:21] The next operator agent is busy processing another workpiece.
[00:00:21] Operator agent calls the operation 'wait(5)'.
[00:00:26] Communication initiated with the next operator to determine the subsequent action.
[00:00:26] The next operator agent is ready.
Output:

*Readers can try this prompt example on various LLMs to test the response.*